%% file: main.tex
\documentclass[11pt, a4paper]{lumia}
\pdfoutput=1

\usepackage[numbers,sort&compress]{natbib}
\usepackage{graphicx}           %
\usepackage{tikz}               %
\usepackage[edges]{forest}      %

\usepackage{url}                %
\usepackage{xurl}               %

\usepackage{array}              %
\usepackage{longtable}          %
\usepackage{multirow}           %
\usepackage{makecell}           %
\usepackage{ragged2e}           %

\usepackage{mathtools}          %
\usepackage{nicefrac}           %

\usepackage{algorithm}          %
\usepackage{algorithmicx}       %
\usepackage{algpseudocode}      %
\usepackage{listings}           %

\usepackage{subcaption}         %
\usepackage{wrapfig}            %
\usepackage[export]{adjustbox}  %

\usepackage{changes}            %
\usepackage{xspace}             %
\usepackage[normalem]{ulem}     %
\usepackage{CJKutf8}            %

\usepackage[tikz]{bclogo}       %
\usepackage[framemethod=tikz]{mdframed} %

\usepackage{placeins}
\usepackage{cleveref}
\usepackage{tocloft}            %
\usepackage{afterpage}          %
\usepackage{bbding}             %
\usepackage{epigraph}           %
\usepackage{minitoc}            %
\usepackage{multicol}           %
\usepackage{textgreek}          %

\definecolor{mygreen}{rgb}{0.29, 0.7, 0.48}
\definecolor{darksalmon}{rgb}{0.91, 0.59, 0.48}
\definecolor{mygrey}{gray}{0.4}

\usepackage[most]{tcolorbox}
\usepackage{makecell}
\newcolumntype{P}[1]{>{\RaggedRight\arraybackslash}p{#1}}

\definecolor{uclablue}{RGB}{39, 116, 174}
\definecolor{bigaired}{RGB}{156, 0, 0}
\definecolor{myblue}{HTML}{598BE7}
\definecolor{mildblue}{RGB}{31,119,180}
\definecolor{sectionblue}{RGB}{70, 130, 180}
\definecolor{methodblue}{RGB}{0, 150, 136}
\definecolor{bgblue}{RGB}{245,243,253}
\definecolor{ttblue}{RGB}{91,194,224}
\definecolor{mygreen}{rgb}{0.64, 0.56, 0.88}
\definecolor{myyellow}{rgb}{0.68, 0.6, 0.1}
\definecolor{fancygreen}{rgb}{0.33, 0.68, 0.20}
\definecolor{salmon}{rgb}{0.94, 0.52, 0.49}
\definecolor{tablegreen}{rgb}{0.82, 0.94, 0.75}
\definecolor{tableblue}{rgb}{0.81, 0.90, 0.94}
\definecolor{tablered}{rgb}{0.97, 0.85, 0.85}
\definecolor{tableorange}{rgb}{0.96, 0.85, 0.81}
\definecolor{myorange}{rgb}{1.0, 0.49, 0.0}
\definecolor{tlgreen}{rgb}{0.33, 0.68, 0.20}
\definecolor{darkgreen}{RGB}{0,100,0}
\definecolor{darkred}{RGB}{200, 0, 0}
\definecolor{customyellow}{HTML}{FFFACD}
\definecolor{refinegreen}{RGB}{0, 128, 75}
\definecolor{scoregreen}{RGB}{34, 139, 34}
\definecolor{hidden-blue}{RGB}{194,232,247}
\definecolor{hidden-black}{RGB}{20,68,106}
\definecolor{yes}{HTML}{C6EFCE}
\definecolor{no}{HTML}{FFC7CE}
\definecolor{partial}{HTML}{FFEB9C}
\definecolor{external}{HTML}{D9E1F2}
\definecolor{hdr}{HTML}{F2F2F2}
\definecolor{GRPOrow}{gray}{0.96}
\definecolor{FlowRLrow}{RGB}{225,236,255}
\definecolor{FlowBlue}{RGB}{80,120,210}
\definecolor{GRPOGray}{gray}{0.35}

\setlist[itemize]{leftmargin=20pt, noitemsep, topsep=0pt}

\newcommand{\cmark}{\textcolor{darkgreen}{\boldmath$\checkmark$}}
\newcommand{\xmark}{\textcolor{darkred}{\boldmath$\times$}}

\newenvironment{itemize*}%
 {\leftmargini=10pt\begin{itemize}%
  \setlength{\itemsep}{0pt}%
  \setlength{\parskip}{0pt}%
  }%
 {\end{itemize}}

\newenvironment{enumerate*}%
 {\begin{enumerate}%
  \setlength{\itemsep}{0pt}%
  \setlength{\parskip}{0pt}}%
 {\end{enumerate}}

\newcommand{\cellstatus}[1]{%
  \begingroup
  \StrTrim{#1}[\statusval]%
  \IfStrEq{\statusval}{Yes}{\cellcolor{yes}\cmark}{}%
  \IfStrEq{\statusval}{No}{\cellcolor{no}\xmark}{}%
  \IfBeginWith{\statusval}{Yes (}{\cellcolor{yes}\cmark~\textit{\statusval\unskip}}{}%
  \IfStrEq{\statusval}{Partial}{\cellcolor{partial}\textbf{Partial}}{}%
  \IfStrEq{\statusval}{External}{\cellcolor{external}\textbf{External}}{}%
  \endgroup
}

\newtcolorbox{myboxi}[1][]{
  breakable,
  title=#1,
  colback=red!5,
  colbacktitle=red!5,
  coltitle=black,
  fonttitle=\bfseries,
  bottomrule=0pt,
  toprule=0pt,
  leftrule=2pt,
  rightrule=2pt,
  titlerule=0pt,
  arc=0pt,
  outer arc=0pt,
  colframe=red,
}

\newtcolorbox{myboxnote}[1][]{
  breakable,
  title=#1,
  colback=orange!0,
  colbacktitle=orange!0,
  coltitle=black,
  fonttitle=\bfseries,
  bottomrule=0pt,
  toprule=0pt,
  leftrule=2pt,
  rightrule=2pt,
  titlerule=0pt,
  arc=0pt,
  outer arc=0pt,
  colframe=orange,
}

\newtcolorbox{myboxii}[1][]{
  breakable,
  freelance,
  title=#1,
  colback=white,
  colbacktitle=white,
  coltitle=black,
  fonttitle=\bfseries,
  bottomrule=0pt,
  boxrule=0pt,
  colframe=white,
  overlay unbroken and first={
  \draw[red!75!black,line width=3pt]
    ([xshift=5pt]frame.north west) -- 
    (frame.north west) -- 
    (frame.south west);
  \draw[red!75!black,line width=3pt]
    ([xshift=-5pt]frame.north east) -- 
    (frame.north east) -- 
    (frame.south east);
  },
  overlay unbroken app={
  \draw[red!75!black,line width=3pt,line cap=rect]
    (frame.south west) -- 
    ([xshift=5pt]frame.south west);
  \draw[red!75!black,line width=3pt,line cap=rect]
    (frame.south east) -- 
    ([xshift=-5pt]frame.south east);
  },
  overlay middle and last={
  \draw[red!75!black,line width=3pt]
    (frame.north west) -- 
    (frame.south west);
  \draw[red!75!black,line width=3pt]
    (frame.north east) -- 
    (frame.south east);
  },
  overlay last app={
  \draw[red!75!black,line width=3pt,line cap=rect]
    (frame.south west) --
    ([xshift=5pt]frame.south west);
  \draw[red!75!black,line width=3pt,line cap=rect]
    (frame.south east) --
    ([xshift=-5pt]frame.south east);
  },
}

\tcbset{
  takeawaysbox/.style={
    title=Takeaways,
    colback=lightblue!80,
    colframe=black,
    fonttitle=\bfseries\small,
    coltitle=white,
    colbacktitle=black,
    enhanced,
    attach boxed title to top left={xshift=2.5mm,yshift=-2.5mm},
    boxed title style={rounded corners, size=small, colframe=black, colback=black},
    width=\linewidth,
    arc=3.5mm
  }
}

\mdfdefinestyle{mystyle}{%
  rightline=true,
  innerleftmargin=10,
  innerrightmargin=10,
  outerlinewidth=3pt,
  topline=false,
  rightline=true,
  bottomline=false,
  skipabove=\topsep,
  skipbelow=\topsep
}

\tikzset{%
    every node/.style={font=\tiny},
    parent/.style =          {align=center,text width=2cm,rounded corners=3pt, line width=0.3mm, fill=gray!10,draw=gray!80},
    child/.style =           {align=center,text width=2.0cm,rounded corners=3pt, fill=blue!10,draw=blue!80,line width=0.3mm},
    grandchild/.style =      {align=center,text width=2cm,rounded corners=3pt},
    greatgrandchild/.style = {align=center,text width=1.5cm,rounded corners=3pt},
    greatgrandchild2/.style = {align=center,text width=1.5cm,rounded corners=3pt},    
    referenceblock/.style =  {align=center,text width=1.5cm,rounded corners=2pt},
    pretrain/.style =           {align=center,text width=2.0cm,rounded corners=3pt, fill=blue!10,draw=blue!80,line width=0.3mm},   
    pretrain_work/.style =           {align=center, text width=8.5cm,rounded corners=3pt, fill=blue!10,draw=blue!0,line width=0.3mm},  
    template/.style =           {align=center,text width=2.0cm,rounded corners=3pt, fill=red!10,draw=red!80,line width=0.3mm},   
    template_work/.style =           {align=center,text width=8.5cm,rounded corners=3pt, fill=red!10,draw=red!0,line width=0.3mm},    
    answer/.style =           {align=center,text width=2.0cm,rounded corners=3pt, fill= cyan!10,draw= cyan!80,line width=0.3mm},   
    answer_work/.style =           {align=center,text width=8.5cm,rounded corners=3pt, fill= cyan!10,draw= cyan!0,line width=0.3mm},      
    multiple/.style =           {align=center,text width=2.0cm,rounded corners=3pt, fill= orange!10,draw= orange!80,line width=0.3mm},   
    multiple_work/.style =           {align=center,text width=8.5cm,rounded corners=3pt, fill= orange!10,draw= orange!0,line width=0.3mm},        
    tuning/.style =           {align=center,text width=2.0cm,rounded corners=3pt, fill= magenta!10,draw= magenta!80,line width=0.3mm},   
    tuning_work/.style =           {align=center,text width=8.5cm,rounded corners=3pt, fill= magenta!10,draw= magenta!0,line width=0.3mm},          
}

\newcommand{\lstbg}[3][0pt]{{\fboxsep#1\colorbox{#2}{\strut #3}}}

\lstdefinelanguage{diff}{
  basicstyle=\ttfamily\small,
  morecomment=[f][\lstbg{red!20}]-,
  morecomment=[f][\lstbg{green!20}]+,
}

\lstdefinelanguage{diffpython}{
  language=diff,
  morekeywords={def, if, else, for, while, return, import, from, as, class, with, try, except, finally, raise, lambda, and, or, not, in, is, None, True, False},
  morecomment=[l]{\#},
  morestring=[b]",
  morestring=[b]',
}

\usepackage{dsfont}
\usepackage{fvextra}
\usepackage{fancyvrb}

\setheadertext{
  \includegraphics[height=1.2cm]{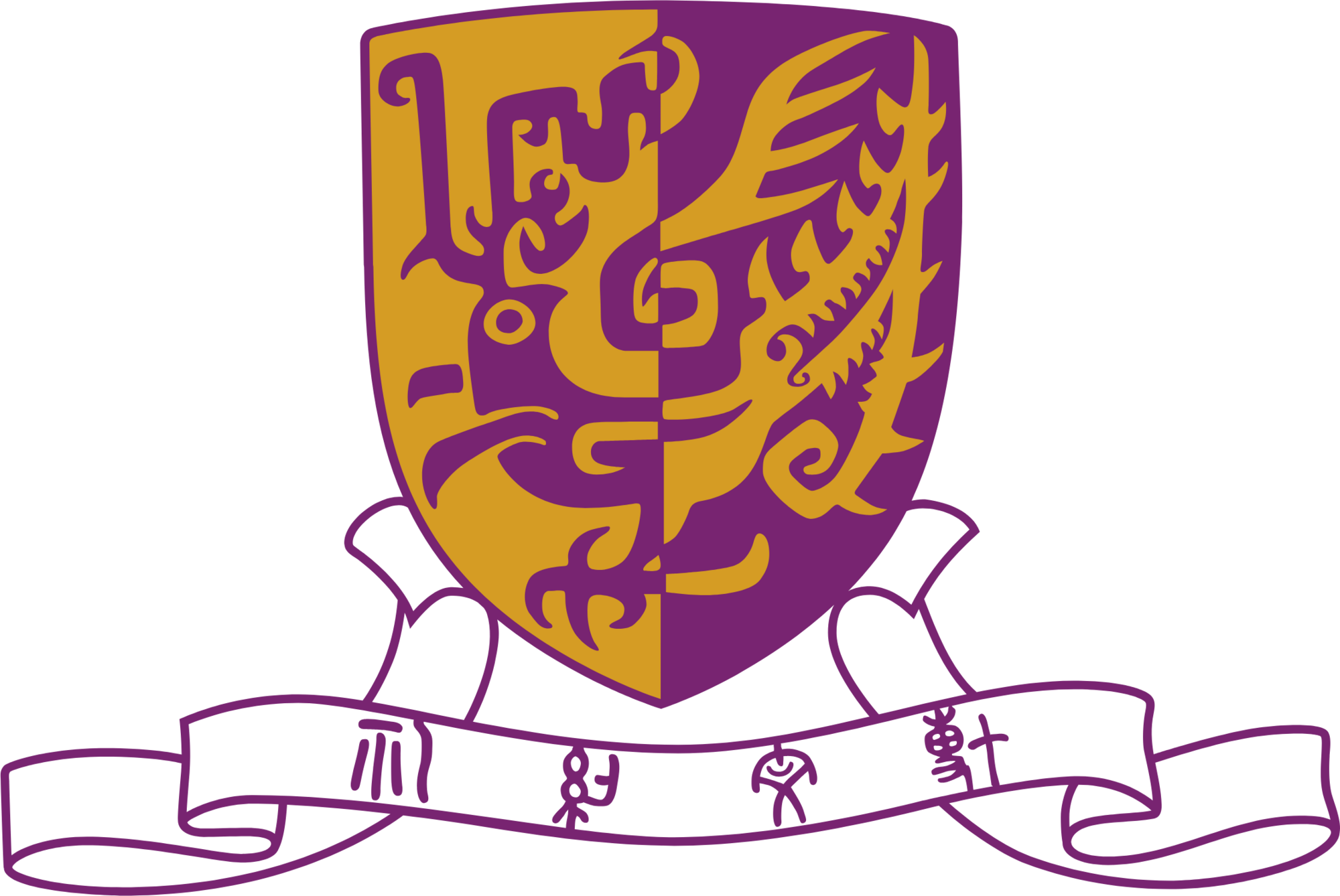} \hspace{0.1cm}
  \includegraphics[height=1.2cm]{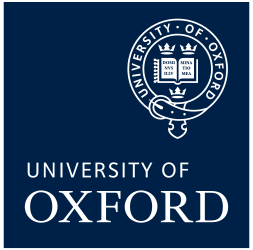} \hspace{0.1cm}
  \includegraphics[height=1.2cm]{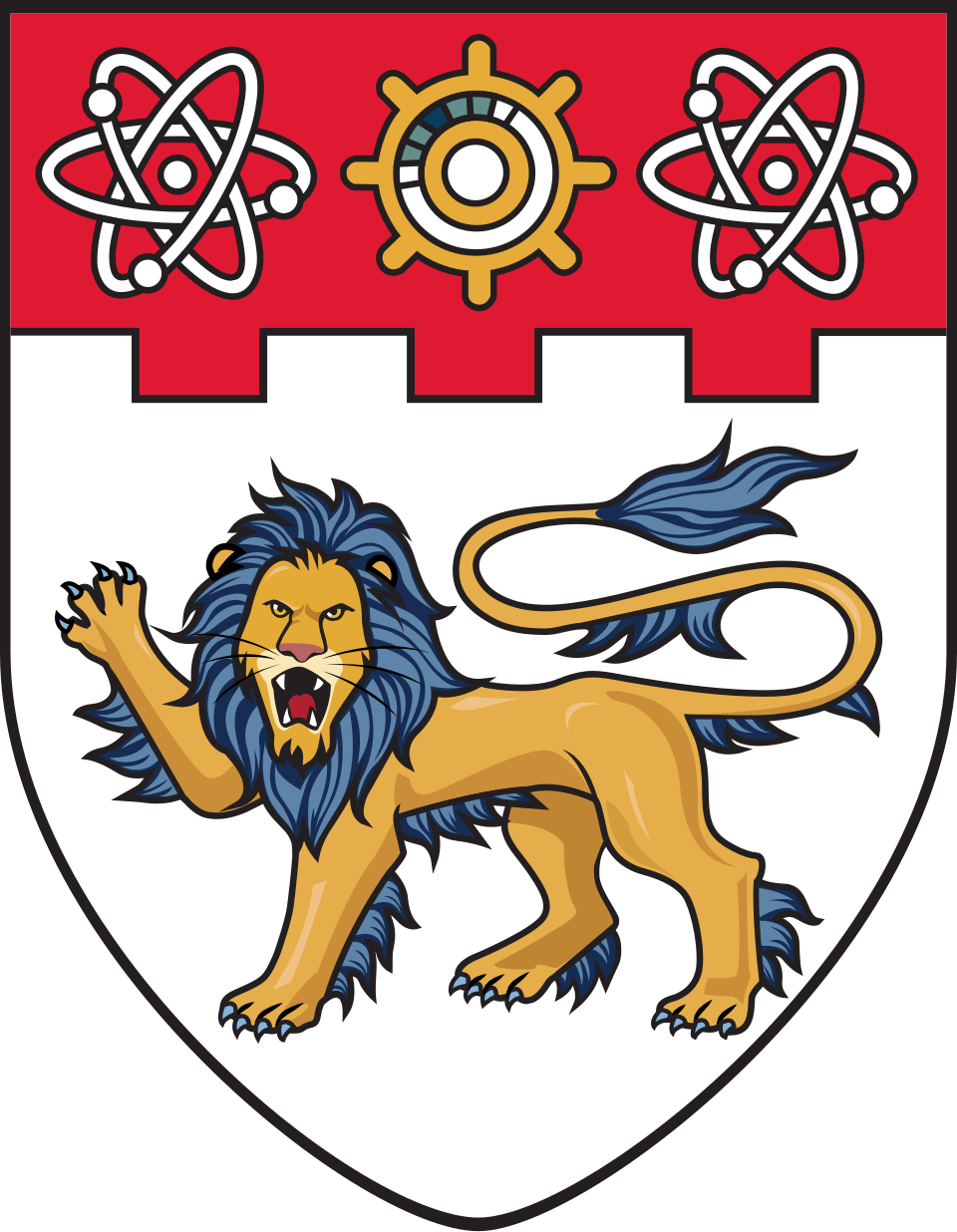} \hspace{0.1cm}
  \includegraphics[height=1.2cm]{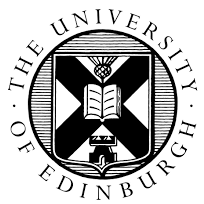} \hspace{0.1cm}
  \includegraphics[height=1.2cm]{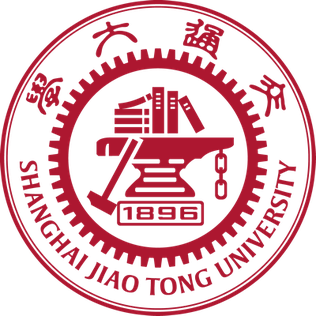}
}

\correspondingemail{
  {\emailicon\ }\href{mailto:dzyu23@cse.cuhk.edu.hk}{dzyu23@cse.cuhk.edu.hk} \quad $^\dagger$Corresponding authors. \newline
}

\renewcommand{\today}{2026-05-20}

\newcommand{\methodname}{MoLEM}
\newcommand{\diffBlue}[1]{\rlap{\ensuremath{^{\color{blue}\scriptscriptstyle +#1}}}}
\newcommand{\diffRed}[1]{\rlap{\ensuremath{^{\color{red}\scriptscriptstyle -#1}}}}
\NewDocumentCommand{\hongru}{ mO{} }{\textcolor{red}{\textsuperscript{\textit{Hongru}}\textsf{\textbf{\small[#1]}}}}

\title{Dynamic Mixture of Latent Memories for Self-Evolving Agents}
\setheadertitle{Dynamic Mixture of Latent Memories for Self-Evolving Agents}

\author{
  Dianzhi Yu$^{1}$, Vireo Zhang$^{3}$, Hongru Wang$^{4}$, Yanyu Chen$^{1}$, Minda Hu$^{1}$, \newline
  Wanghan Xu$^{5}$, Siki Chen$^{2}$, Philip Torr$^{2}$, Zhenfei Yin$^{2\dagger}$, Irwin King$^{1\dagger}$ \\
  $^1$The Chinese University of Hong Kong \quad
  $^2$University of Oxford \quad
  $^3$Nanyang Technological University \newline
  $^4$University of Edinburgh \quad
  $^5$Shanghai Jiao Tong University
}

\begin{document}

\input{Content/00_abstract}

\maketitle

\input{Content/01_introduction}
\input{Content/02_related_work}
\input{Content/03_preliminary}
\input{Content/04_methodology}
\input{Content/05_experiments}
\input{Content/06_conclusion}

\bibliography{MyLibrary,ref}

\appendix

\input{Content/07_appendix}

\end{document}

%% file: Content/00_abstract.tex
\begin{abstract}

Achieving self-evolution in intelligent agents requires the continual accumulation of new knowledge across changing task sequences without forgetting previously acquired abilities.
Existing approaches either internalize knowledge by updating model parameters, which induces catastrophic forgetting, or rely on external memory, which fails to genuinely enhance the model's intrinsic capabilities.
We propose \textbf{\methodname}, a generative mixture of latent memory framework based on a dynamic mixture-of-experts (MoE).
We treat multiple experts as independent carriers to generate memory. A router selects and weights experts through key-query matching, and the aggregated latent memory is injected into the reasoning process.
The base model for reasoning remains entirely frozen, with all experiential knowledge internalized into the additional modules, avoiding catastrophic forgetting.
For continual learning, each training stage is paired with a lightweight autoencoder that selects the appropriate routing group at inference, and inputs that match no stage fall back to the pretrained model.
Experiments train the framework on continual-learning sequences spanning math, science, and code domains.
After training, we evaluate the framework on the corresponding test sets to measure task learning and competence preservation across continual adaptation stages.
After the full continual-learning sequence, our method improves the average accuracy by $10.40\%$ over the Vanilla pretrained baseline, while none of the competing methods consistently exceed this baseline across different training orders.

\end{abstract}

%% file: Content/01_introduction.tex
\section{Introduction}

Large Language Model (LLM) agents have achieved significant progress within individual domains~\citep{Gao2025Survey,Zhang2025MemGena}.
However, most existing work focuses on optimizing such per-domain performance in isolation, and it remains unclear how to build a genuinely versatile agent that continually evolves across a sequence of diverse tasks.
Achieving this requires the agent's \textbf{self-evolving} capability. Agents need to acquire new knowledge through parameter and memory updates across continually changing task sequences, while avoiding forgetting and preserving general-purpose competence.
Thus, our work aims to enable agents to internalize experience through agent memory, integrate new domain knowledge without erasing prior competence, and progressively enhance problem-solving capability across continually evolving task sequences~\citep{Gao2025Survey}.

Existing memory mechanisms for agents generally fall into two categories. 
The first is \textbf{parametric memory}, which learns knowledge by updating the agent's model parameters (e.g., via SFT or GRPO~\citep{Shao2024DeepSeekMath}).
Although modifying model parameters improves performance, it inevitably causes \textit{catastrophic forgetting}, disrupting performance on previously trained downstream tasks~\citep{McCloskey1989Catastrophic,Ratcliff1990Connectionist} and degrading pretrained capabilities~\citep{Zheng2023Preventing,Yu2026Recent,Hassabis2017NeuroscienceInspired}.
This approach also causes interference between tasks due to shared parameters~\citep{Wang2024Comprehensive}. 
In contrast, the second category \textbf{retrieval-based memory} saves past experience into external structured databases.
While this approach avoids catastrophic forgetting since model parameters are not changed, it remains a form of context engineering~\citep{Zhang2025MemGena}. It relies on retrieved contexts to elicit the model's preexisting capabilities, yet the model itself acquires no new abilities.
External databases remain disconnected from the model's internal weights, which prevents genuine learning.
Both approaches have distinct drawbacks.
An ideal memory mechanism should, without intruding on the original parameters, internalize experience into latent representations that the model can directly leverage, which motivates \textbf{latent memory} approaches.

Recent latent-memory methods, including MemGen~\citep{Zhang2025MemGena}, SoftCoT~\citep{Xu2025SoftCoT}, and Coconut~\citep{Hao2025Training}, have made progress along this direction by leveraging latent states as high-density memory carriers.
These methods show that continuous internal representations can mediate between parameter updates and external retrieval.
However, for agent self-evolution, existing latent memory methods still suffer from three key limitations:
\textbf{(1) Static architecture:} existing frameworks have a fixed model structure and cannot dynamically extend their capacity in response to task difficulty or new domains;
\textbf{(2) Forgetting in learned latent memory:} although latent memory avoids updating the backbone, it is still produced by trainable parameters; naively reusing the same memory module across tasks can overwrite previously learned memory behavior;
\textbf{(3) Inflexible module separation:} existing frameworks often separate different modules for specified functions and execute fixed pipelines, lacking the flexibility to integrate multiple cognitive parts in the way the human brain does.
Therefore, a critical research question remains:

\textit{How can we design a flexible, extensible cognitive architecture for agents such that, when faced with continuously changing tasks, they can both dynamically expand capacity to acquire new knowledge and avoid forgetting previously learned abilities?}

To address this challenge, we propose \textbf{\methodname}, a generative latent memory framework built upon a dynamic mixture-of-experts (MoE).
Multiple experts act as separate memory carriers; a router selects and weights relevant experts through key-query matching, and the aggregated latent memory is injected into the reasoning process.
Each stage's router is paired with a lightweight autoencoder: at inference, the routing group with the smallest reconstruction error is selected, and inputs that match no stage fall back to the pretrained model.
By keeping the base model for reasoning frozen and updating only the MoE memory modules, our framework internalizes new experience without overwriting the pretrained reasoning backbone.

\textbf{Experimental Validation.}
We conduct extensive experiments under continual learning (CL) settings, with task sequences spanning math, science, and code domains.
After each training stage, we evaluate the framework on the corresponding test sets to assess new-task learning and competence preservation along the continual-learning sequence.
After the full continual-learning sequence, our method improves the average accuracy by $10.40\%$ over the Vanilla pretrained baseline, while none of the competing methods consistently exceed this baseline across different training orders.

In summary, the main contributions of this paper are as follows:
\begin{itemize}
\item We propose a dynamic mixture of latent memory framework that expands memory capacity through multiple experts and uses key-query routing to select and weight relevant experts for latent-memory generation;
\item We introduce a task-ID-free domain-aware routing mechanism for domain-incremental settings: stage-wise autoencoders select the most compatible routing group by reconstruction error, while inputs that match no stage fall back to the pretrained model for OOD protection;
\item We conduct systematic experiments on continual-learning task sequences spanning math, science, and code, evaluating the corresponding three test sets after each training stage to assess new-task learning and competence preservation.
\end{itemize}

%% file: Content/02_related_work.tex
\section{Related Work}

\paragraph{Parametric Methods.}
Directly fine-tuning LLM parameters via methods such as SFT, GRPO~\citep{Shao2024DeepSeekMath}, and DPO~\citep{Rafailov2024Direct} is the most direct route to internalizing task knowledge.
Several works including RLVMR~\citep{Zhang2025RLVMR}, MEM1~\citep{Zhou2025MEM1}, and Memory-R1~\citep{Yan2025MemoryR1} further integrate reasoning steps or memory mechanisms into the training loop, but they still rely on parametric training to update model behavior.
Among these, RLVMR uses meta-reasoning labels to annotate intermediate reasoning steps, focusing on process rather than only the final outcome;
MEM1 compresses contextual information into an internal state representation that supports both memory consolidation and reasoning;
Memory-R1 leverages reinforcement learning to jointly optimize the agent's memory management policy.
Although these methods introduce reasoning steps or memory mechanisms, they still absorb experience through parameter updates, and therefore remain exposed to catastrophic forgetting under continual-learning settings.

\paragraph{Retrieval-based Memory.}
MemoryBank~\citep{Zhong2024Memorybank} introduces the Ebbinghaus forgetting curve to dynamically update memory strength, while Memento~\citep{Zhou2025Memento} bypasses LLM parameter modification entirely and maintains an external memory system via case-based reasoning.
ExpeL~\citep{Zhao2024ExpeL} further demonstrates the value of experience-based external memory: it extracts natural-language insights from successful and failed trajectories and, at test time, retrieves both similar historical experiences and abstract lessons to assist decision-making.
A-Mem~\citep{Xu2025AMEM} dynamically organizes memories by constructing structured notes, establishing inter-memory links, and evolving existing memories based on new experiences, while ReasoningBank~\citep{Ouyang2025ReasoningBank} extracts structured memory entries and dynamically leverages past trajectories to improve reasoning through test-time learning.
While these non-invasive methods avoid catastrophic forgetting, they remain forms of context engineering: the model's intrinsic capabilities are not improved.

\paragraph{Latent Memory.}
MemGen~\citep{Zhang2025MemGena} introduces a latent-memory framework that weaves latent states into the reasoning process: its memory trigger monitors the reasoning state to decide when to invoke memory, while its memory weaver constructs latent token sequences from the current state.
SoftCoT~\citep{Xu2025SoftCoT} employs an additional small auxiliary model to generate continuous reasoning tokens that are injected into the original LLM via a projection module, while Coconut~\citep{Hao2025Training} directly reuses hidden states as input embeddings to realize a continuous chain of thought.
Latent memory uses continuous vectors as high-density carriers of memory, enabling knowledge internalization without modifying the original model parameters.
However, existing latent-memory methods do not consider architectural scalability under continual-learning settings, which is precisely the central concern of this work.

%% file: Content/03_preliminary.tex
\section{Preliminary}

\subsection{Latent Memory Generation}

Latent-memory generation uses learned modules to produce a continuous latent sequence during reasoning, and injects this sequence as an auxiliary internal representation to support later reasoning and next-token prediction~\citep{Zhang2025MemGena}.
Our framework organizes this mechanism with a dynamic MoE architecture: different experts generate different latent memories, a router selects the relevant experts, and the resulting latent memories are aggregated into a more flexible auxiliary signal for reasoning.
We use a delimiter token set $\mathcal{D}$ (e.g., commas and periods) where latent-memory generation can be activated~\citep{Zhang2025MemGena}.

\subsection{Notation}

Let $\pi_\theta$ denote the pretrained reasoner, whose parameters $\theta$ remain frozen throughout the framework.
The MoE module consists of $K$ experts $\{E_1, \ldots, E_K\}$, each an independent adapter parameterized by $\phi_k$ and attached to $\pi_\theta$.
The router $R_\psi$ (parameterized by $\psi$) is similarly an adapter attached to $\pi_\theta$, and makes routing decisions at every delimiter position.
At the $t$-th delimiter position, we distinguish three hidden-state signals: $\mathbf{H}_t$ produced by the base reasoner alone, $\mathbf{H}_t^{E_k}$ produced when expert $k$'s adapter is active, and $\mathbf{H}_t^{R}$ produced when the router's adapter is active.
Each expert generates a candidate latent memory $\mathbf{M}_{k,t}$ from $\mathbf{H}_t^{E_k}$, the router assigns expert weights $\alpha_{k,t}$ from $\mathbf{H}_t^{R}$, and we use $\mathbf{M}_t$ to denote the final aggregated latent memory injected into the reasoner.

\subsection{Problem Formulation}

We consider the \textbf{domain-incremental learning} setting~\citep{vandeVen2022Three}  in its more realistic, \emph{task-ID-agnostic} form: the task identity is not provided at inference time, and the model must operate without an external task label.
Given a task sequence over $T$ domains $\mathcal{D}_1, \mathcal{D}_2, \ldots, \mathcal{D}_T$ (e.g., math, science, code), the framework only accesses $\mathcal{D}_t$'s training data during phase $t$.
The objective is to maintain or improve performance on all previously seen domains:
\begin{equation}
\max_{\{\phi_k\}, \psi} \; \frac{1}{T} \sum_{t=1}^{T} \text{Acc}(\pi_\theta, \{E_k\}, R_\psi; \mathcal{D}_t)
\end{equation}
where $\theta$ is held fixed and optimization is performed only over the expert parameters $\{\phi_k\}$ and the router parameters $\psi$.

Remark of Competence Preservation under Self-Evolving Adaptation:
Since our framework uses a pretrained reasoner $\pi_\theta$ whose parameters remain frozen, continual adaptation should not degrade the general reasoning capabilities already acquired during pretraining.
Let $\mathrm{Acc}_0(\mathcal{D})$ denote the accuracy of the frozen pretrained reasoner on dataset $\mathcal{D}$ without any learned memory module.
This quantity serves as the competence baseline that adaptation should preserve on domains not covered by the current training stage.
We therefore evaluate the model on the stage-wise test sets $\{\mathcal{D}_u^{\mathrm{test}}\}_{u=1}^{T}$ throughout the continual-learning sequence, focusing on two goals: on learned domains, the system should absorb new knowledge while preserving previous performance; on out-of-distribution (OOD) domains that have not yet been covered by adaptation, performance should remain close to the pretrained competence baseline:
\begin{equation}
\mathrm{Acc}(\pi_\theta, \{E_k\}, R_\psi;\, \mathcal{D}^{\mathrm{ood}}) \gtrsim \mathrm{Acc}_0(\mathcal{D}^{\mathrm{ood}}).
\end{equation}

%% file: Content/04_methodology.tex
\section{Methodology}

\subsection{Overview: Dynamic MoE with Latent Memory}

\begin{figure}[t]
\centering
\includegraphics[width=\linewidth]{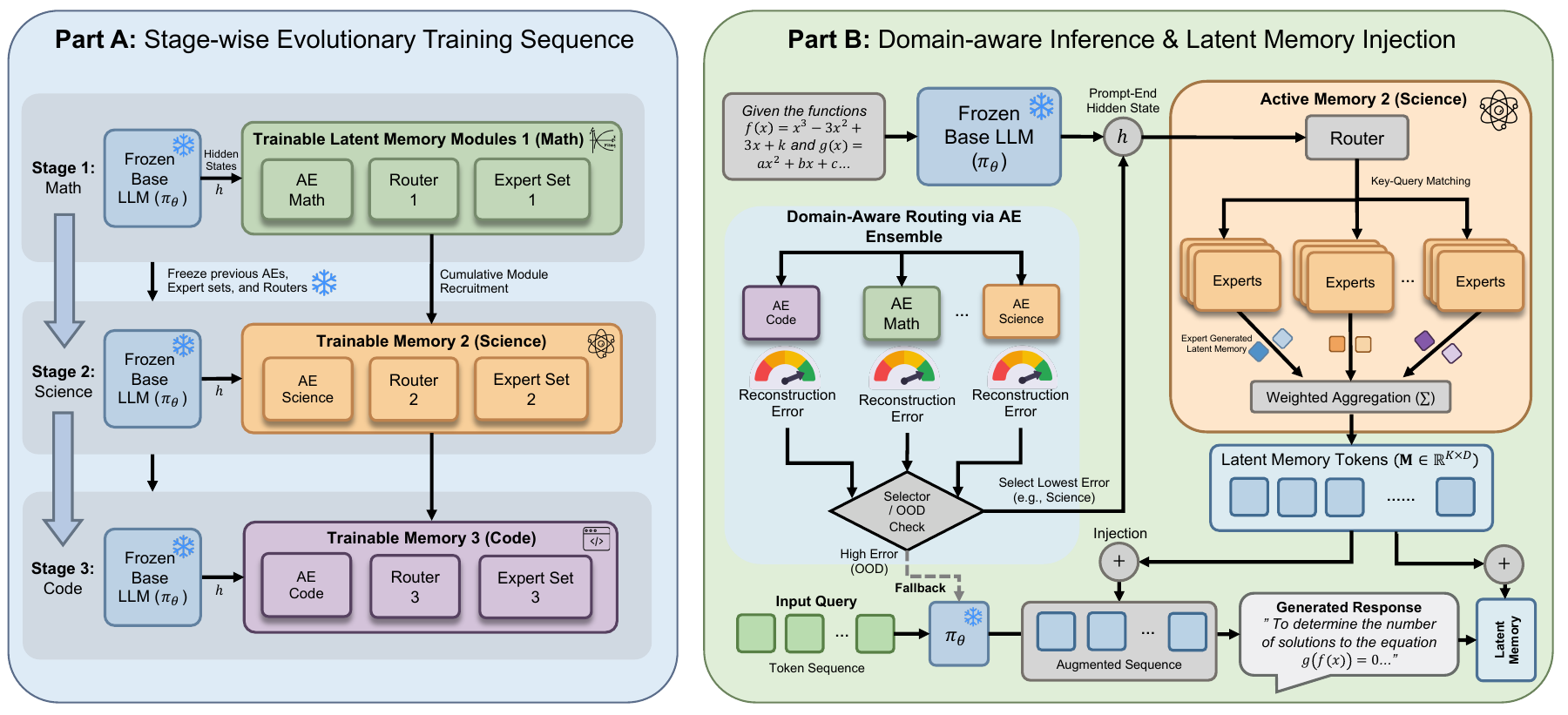}
\caption{\textbf{Overview of \methodname.}
Each expert is attached to the frozen base reasoner and produces a candidate latent memory.
The router performs key-query matching to compute expert weights and aggregates the candidates into the final latent memory that is injected back into the reasoner.
For continual learning, each stage is paired with a lightweight autoencoder (AE); at inference, the AE with the smallest reconstruction error selects the corresponding routing group, while inputs whose reconstruction errors all exceed their thresholds are classified as out-of-distribution and fall back to the pretrained model.}
\label{fig:overview}
\end{figure}

As illustrated in Figure~\ref{fig:overview}, we propose a dynamic MoE latent-memory framework built around three core components.
\textbf{(1) Experts:} each expert is an independent module attached to the base reasoner, producing its own distinct latent memory; the outputs of multiple experts are combined by weighted averaging.
\textbf{(2) Router:} the router is another module attached to the same base reasoner; it performs key-query matching to select relevant experts and computes the weights used to aggregate their latent memories.
\textbf{(3) Per-stage autoencoder (AE):} each stage's router is paired with a lightweight AE trained to reconstruct that stage's prompt-end feature distribution; at inference, the routing group with the smallest reconstruction error is selected, and inputs whose errors all exceed a threshold are classified as out-of-distribution (OOD) and fall back to the pretrained model (details in Section~\ref{sec:ae_routing}).
Each selected expert generates a fixed-size latent memory matrix that is weighted aggregated and injected back into the reasoning states.

Our setting is \textbf{domain-incremental learning}: the model has no access to the current task ID at inference time (task-ID agnostic).
Crucially, because the base reasoner has been pretrained on diverse data, its hidden-state representations naturally carry discriminative information across domains; we exploit this inherent discriminability to select the appropriate routing group and to identify out-of-distribution inputs, without relying on any explicit domain label.
The routing mechanism is described in Section~\ref{sec:ae_routing}, and its empirical analysis is presented in Section~\ref{sec:analysis}.
The core reasoner is kept frozen throughout training and inference, and parameter updates occur only within the MoE memory module, avoiding overwrites of the pretrained reasoning backbone and reducing cross-domain parameter interference.

\subsection{Latent-Memory Experts}

Each expert $E_k$ is realized as an independent LoRA adapter~\citep{Hu2022LoRA} attached to the frozen base reasoner $\pi_\theta$.
At an invocation step $t$, the context is forwarded through $\pi_\theta$ with expert $E_k$ active, producing an expert-side hidden state sequence $\mathbf{H}_t^{E_k}$.
The expert then generates latent memory matrix $\mathbf{M}_{k,t}$ based on this, which is injected back into the reasoning hidden states to influence subsequent generation~\citep{Zhang2025MemGena}.
Because the experts are parameter-isolated, the same input gives rise to different latent memories under different experts; this isolation underlies the per-domain specialization that we exploit in the continual-learning setting.

\subsection{Key-query routing}

\paragraph{Query extraction.}
The router is a LoRA adapter attached to the same base reasoner $\pi_\theta$ as the experts.
At delimiter position $t$, the context is forwarded through $\pi_\theta$ with the router active, producing a router-modified hidden state sequence $\mathbf{H}_t^{R}$.
We take the hidden state at its last position, denoted $\mathbf{h}_t^{R}$, and apply a lightweight projection to obtain the query vector $\mathbf{q}_t = \mathrm{Proj}(\mathbf{h}_t^{R})$.

\paragraph{Key-query matching.}
Each expert $k$ maintains a learnable key vector $\mathbf{k}_k$, and the router computes a matching score by cosine similarity:
\begin{equation}
s_{k,t} = \cos(\mathbf{q}_t, \mathbf{k}_k) = \frac{\mathbf{q}_t \cdot \mathbf{k}_k}{\|\mathbf{q}_t\| \|\mathbf{k}_k\|}.
\end{equation}
The query represents the current reasoning state while each key serves as a learnable routing prototype for one expert; the resulting scores yield expert weights $\alpha_{k,t}$ used to select and aggregate experts.

\subsection{Training Objective}

We train the MoE memory module with supervised fine-tuning while $\pi_{\theta}$ stays frozen.
Let $\mathcal{H}=\{(\mathbf{x}_i,\mathbf{z}_i^*)\}_{i=1}^{N}$ be the training samples, with target token sequence $\mathbf{z}_i^*=(z_{i,1}^*,\ldots,z_{i,L_i}^*)$.
At each invocation step $j$, the experts produce candidate memories $\mathbf{M}_{k,i,j}$ from their expert-modified hidden states $\mathbf{H}_{i,<j}^{E_k}$, and the router aggregates these candidates as
\begin{equation}
\mathbf{M}_{i,j} = \sum_{k=1}^{K} \alpha_{k,i,j}\mathbf{M}_{k,i,j}.
\end{equation}
The SFT objective then maximizes, for every target token, the probability the frozen reasoner assigns to it given the previous tokens and this aggregated memory~\citep{Zhang2025MemGena}:
\begin{equation}
\mathcal{L}_{\text{SFT}}(\phi)
= - \mathbb{E}_{(\mathbf{x}_i,\mathbf{z}_i^*) \sim \mathcal{H}}
\left[
\sum_{j=1}^{L_i}
\log \pi_{\theta}\!\left(z_{i,j}^{*}\mid \mathbf{x}_i, \mathbf{z}_{i,<j}^{*}, \mathbf{M}_{i,j}\right)
\right],
\end{equation}
Here $\phi$ denotes the parameters of the MoE memory module, including the experts and router, while $\theta$ remains fixed.
We further apply a \textbf{load-balance loss} on expert selection to prevent expert usage from concentrating excessively on a small subset of experts during training.
In our setting, the balancing statistics are computed over \emph{routing instances}: each instance corresponds to one MoE invocation at a latent-memory insertion position for one training example, rather than to every next-token prediction position.
Since a sequence contains multiple insertion positions, the loss is evaluated separately for each MoE invocation over the mini-batch and accumulated across invocations in the forward pass.
For the trainable expert set $\mathcal{A}$, let $f_k$ denote the fraction of routing instances whose primary selected expert is $k$, and let $p_k$ denote the mean router probability assigned to expert $k$ in $\mathcal{A}$.
We use a Switch-style balancing term~\citep{Fedus2022Switch}:
\begin{equation}
\mathcal{L}_{\text{LB}}
= |\mathcal{A}| \sum_{k \in \mathcal{A}} f_k p_k .
\end{equation}
In the continual-learning setting, experts from previous stages are excluded from $\mathcal{A}$, so the balancing pressure is applied only among trainable experts.
The full training objective combines the SFT loss with this auxiliary term:
\begin{equation}
\mathcal{L} = \mathcal{L}_{\text{SFT}} + \lambda \mathcal{L}_{\text{LB}} .
\end{equation}

\subsection{Continual Expansion: Stage-wise Recruiting and Domain-Aware Routing}

\paragraph{Stage-wise expert recruiting.}
We adopt a \textbf{Stage-wise Recruiting} strategy: when continual learning transitions from domain $\mathcal{D}_t$ to $\mathcal{D}_{t+1}$, the framework executes the following steps:
\begin{enumerate}
\item %
\textbf{Freeze} the entire set of expert and router parameters trained in phase $t$, thereby consolidating the parametric memory acquired for that domain;
\item %
\textbf{Recruit} a fresh group of experts and a corresponding router for phase $t{+}1$, with parameters randomly initialized;
\item %
Only the newly recruited group is updated during phase $t{+}1$ training; backpropagation does not touch any frozen group from previous phases;
\item %
At inference time, all $T$ groups of (router, experts) are candidates, and the domain-identification mechanism described below decides which group to invoke.
\end{enumerate}
This design turns ``when to expand'' into a fully deterministic decision: a new task triggers expansion immediately, without relying on any metric threshold. Moreover, knowledge from different domains is strictly isolated into separate parameter subspaces, avoiding cross-domain parameter interference.

\paragraph{Domain-Aware Routing via Autoencoder.}\label{sec:ae_routing}
After stage-wise recruiting, the system accumulates $T$ independently trained (router, experts) groups; the routing itself is physically separated across stages, with each group dedicated solely to its own domain.
The key inference-time question is how to choose the appropriate routing group without a task label.
We equip each stage with a lightweight autoencoder (AE) that is trained independently from the base LLM.
For each prompt, we first run the frozen reasoner once and take the last-layer hidden state at the final prompt token, namely the prompt-end feature at the latent-memory insertion position, as the AE input feature $\mathbf{h}$.
Each stage AE is a small model trained only to reconstruct this feature by minimizing reconstruction loss.
At inference, the feature $\mathbf{h}$ is compared only against the available AEs associated with stages that have already been recruited in the continual-learning process.
Among these seen-stage AEs, the stage with the smallest reconstruction error is selected for expert routing when it is accepted by the corresponding threshold; otherwise the input is rejected as OOD:
\begin{equation}
s^* = \arg\min_{s\in \mathcal{S}_{\mathrm{seen}}} \ell_s(\mathbf{h}).
\end{equation}
At phase $t$, $\mathcal{S}_{\mathrm{seen}}=\{1,\ldots,t\}$. Each stage has its own rejection threshold $\delta_t$, calibrated on that stage's validation set as the reconstruction-error percentile covering 95\% of in-distribution validation samples.
If the reconstruction errors exceed their validation-calibrated thresholds for all stages, the input is classified as \textbf{OOD}: no expert from any stage is activated, and inference falls back entirely to the pretrained model.
This design explicitly preserves the model's original capabilities on unseen domains, preventing the use of potentially harmful expert memories for inputs that lie outside the training distribution.

\subsection{Continual Learning Pipeline}

We design a continual-learning task sequence spanning three domains: $\text{Math} \rightarrow \text{Science} \rightarrow \text{Code}$.
At each phase, the framework trains its MoE module exclusively on the data of the current domain (the reasoner is always frozen), and evaluates on all seen and unseen domains in order to comprehensively measure forward transfer and catastrophic forgetting.

Given the prompt-end feature from the frozen reasoner, the framework evaluates reconstruction errors only on the stage-wise autoencoders of previously learned stages. If at least one stage is accepted by its stage-specific threshold, we select the router group of the accepted stage with the smallest reconstruction error; otherwise, the input is rejected as OOD. The selected router then performs expert selection through key-query matching.
This task-agnostic design enables routing without requiring any task identifier.

%% file: Content/05_experiments.tex
\section{Experiments}

\subsection{Experimental Setup}

\paragraph{Datasets.}
We design a continual-learning sequence spanning three domains.
\textbf{Training data:}
(1) \textit{Math}, drawn from Nemotron MATH~\citep{NVIDIA2025NVIDIA}, with mathematical reasoning questions;
(2) \textit{Science}, drawn from Nemotron Science~\citep{NemotronPostTrainingDatasetV1}, consisting of multiple-choice scientific questions;
(3) \textit{Code}, drawn from KodCode~\citep{Xu2025Kodcode}, focused on code-generation problems.

\paragraph{Baselines.}
(1) \textbf{Vanilla}: the pretrained model used directly without any fine-tuning;
(2) \textbf{SFT}: sequentially fine-tuning all parameters on the task sequence;
(3) \textbf{MemGen}~\citep{Zhang2025MemGena}: a representative latent-memory baseline;
(4) \textbf{ExpeL}~\citep{Zhao2024ExpeL}: an external-memory method.
\paragraph{Implementation Details.}
We use Qwen3-4B-Instruct-2507~\citep{qwen3technicalreport} as the core reasoner, kept frozen throughout all experiments.
The CL training sequence is $\text{Math} \rightarrow \text{Science} \rightarrow \text{Code}$.
All methods use the same base model and are sequentially trained on the above default order, while the other two task orders are reported in Appendix~\ref{app:order-sensitivity}.
After each training stage, we report per-dataset accuracy on the three test sets, together with the average across them and the CL auxiliary metrics.
Additional AE, router, latent-memory, optimizer, and training hyperparameters are reported in Appendix~\ref{app:implementation-details}.

\paragraph{Metrics.}
We use accuracy as the evaluation metric on each individual test set.
For the continual-learning setting, we use the \textit{Average Performance} across all test sets as the main metric.
We additionally report \textit{Forgetting}~\citep{Chaudhry2018Riemannian}, \textit{Backward Transfer} (BWT), and \textit{Forward Transfer} (FWT)~\citep{Lopez-Paz2017Gradient} as auxiliary metrics for reference.

\subsection{Main Results: Continual Learning Performance}

\input{tables/big_table_3iid_default_with_cl_metrics}

Table~\ref{tab:main_3iid_with_cl_metrics} reports the evaluation results after each phase of training in the CL sequence under the default order.
Appendix~\ref{app:order-sensitivity} further reports the other two task orders.

After the full training sequence (Stage~3), our method reaches an average accuracy of $73.93\%$ across the three test sets, exceeding SFT ($34.40\%$) by $39.53\%$, the strongest competing method MemGen ($63.47\%$) by $10.46\%$, and the Vanilla pretrained reference ($63.53\%$) by $10.40\%$.
On the per-task results, our method attains the highest accuracy on both Math ($59.20\%$) and Code ($74.20\%$) at the final stage, the best per-task performance among all methods.
Forgetting, backward transfer, and forward transfer all remain at zero for our method at every stage where these metrics are defined (i.e., from Stage~2 onward), reflecting the effectiveness of the routing and expert-isolation design: training a new stage neither degrades performance on previously learned domains nor alters the latent memory produced by earlier experts. These auxiliary metrics are reported for reference and should be interpreted alongside absolute performance, since they reflect an inherent trade-off between accuracy and resistance to forgetting~\citep{Huang2023ConSlide,Lopez-Paz2017Gradient,Chaudhry2018Riemannian}.

We further compare each method against the Vanilla pretrained baseline at the end of the CL sequence.
Under the default training order, cross-task interference drives every competing method's Stage~3 Average to around or below the Vanilla reference of $63.53\%$: SFT lands at $-29.13\%$, ExpeL at $-2.06\%$, and MemGen at $-0.06\%$; sequentially training these methods through three domains therefore yields no net improvement over the pretrained model itself.
Among the methods we compare, only ours delivers a positive net gain over the Vanilla baseline after the full sequence, with $+10.40\%$ on Average; this advantage holds consistently across all three training orders examined in Appendix~\ref{app:order-sensitivity}.
This effect is not visible from a single-domain training-and-evaluation setting; it only surfaces once the model has been carried through a multi-stage continual-learning sequence, which is more realistic for self-evolving agents, where parametric updates from later domains erode the competence acquired earlier for other methods.

Besides the Average metric, we also analyse performance for each task.
Baselines with global parameter updates exhibit pronounced degradation on earlier domains as training progresses: SFT accuracy on Math collapses from $57.20\%$ at Stage~1 to $12.80\%$ at Stage~3, and SFT Science drops from $84.20\%$ to $27.20\%$ over the same span; MemGen Math also drops from $58.20\%$ to $32.00\%$ between Stage~1 and Stage~3.
ExpeL, a retrieval-based method that does not update the reasoner's parameters, avoids such parametric drift but exhibits two complementary limitations.
First, on domains that require task-specific learning, such as Math, its accuracy is bounded by the pretrained model's reasoning capability and remains in the $28.00\%$--$32.00\%$ range across all stages, far below the training-based methods (e.g., $59.20\%$ for our method).
Second, retrieval-injected context does not necessarily improve performance: on Code, ExpeL settles at $60.60\%$ at Stage~3, $11.40\%$ below the Vanilla pretrained reference ($72.00\%$), indicating that retrieved exemplars can interfere with code-generation prompts rather than help them.
Taken together, these patterns show that our method preserves earlier-domain capability that neither purely parametric updates nor purely retrieval-based augmentation are able to retain.

\subsection{Ablation Study}

\input{tables/big_table_3iid_ablation_with_cl_metrics}

\paragraph{Multiple experts vs.\ a single expert.}
We compare configurations with and without multiple experts under a single shared routing group, corresponding to the $\times$/$\times$ and $\times$/\checkmark{} rows of Table~\ref{tab:ablation_3iid_with_cl_metrics}.
After Stage~1 on Math, switching from a single expert to multiple experts improves Math accuracy from 53.80 to 59.80, indicating that the additional memory capacity is beneficial even on a single dataset.
This benefit, however, does not transfer to the continual-learning setting when the routing group remains shared across stages: parameter sharing still drives cross-stage interference, and the shared multi-expert variant reaches Forget = 30.50 by Stage~3.
This observation motivates separating routing groups across stages, which we examine next.

\paragraph{Isolated routing groups across stages.}
With Isolation R\&E enabled, each stage owns its own router and experts, so that the parameters trained for earlier stages are not overwritten when later stages are added (the $\times$/\checkmark{} versus \checkmark{}/\checkmark{} rows, and the $\times$/$\times$ versus \checkmark{}/$\times$ rows of Table~\ref{tab:ablation_3iid_with_cl_metrics}).
With multiple experts, switching from shared to isolated routing raises the Stage~3 average from 51.80 to 73.93 and reduces Forget from 30.50 to 0.00, identifying isolation as the dominant factor that suppresses cross-stage interference.
A single expert under isolation already removes most of the forgetting (Stage~3 Forget = 1.10 with average 70.20), yet it cannot match the per-stage capacity of multiple experts: replacing the single-expert isolated variant with the multi-expert isolated configuration further improves Math accuracy from 53.80 to 59.20 and the Stage~3 average from 70.20 to 73.93.
The two design choices are complementary, and combining isolated routing with multiple experts attains the best accuracy together with the lowest forgetting in our continual-learning sequence.

\paragraph{Load-balance loss.}
The load-balance loss encourages more balanced exposure across experts during training, preventing the router from repeatedly concentrating updates on only a small subset of experts.
We probe its effect at Stage~1 Math.
Under this setting, removing the load-balance loss reduces Math Test accuracy from $59.80\%$ to $58.00\%$ under the same configuration, suggesting that the auxiliary objective contributes to effective expert training.
\subsection{Analysis}

\paragraph{Expert specialization analysis.}

Figure~\ref{fig:expert_usage_3iid_overall} shows the overall expert usage on the Math, Science, and Code test sets.
Each stage owns its expert pool, and the figure reports usage for each domain.
During training, the load-balance objective gives experts more balanced exposure and helps each expert receive sufficient updates.
At inference time, however, the router still exhibits input-dependent expert preferences.
Math relies more on E1 and E2, whereas Code distributes usage more broadly while still assigning the largest share to E3.
This pattern suggests spontaneous expert diversity: although no expert role is manually specified, different experts are used with different tendencies across input domains.

The diversity of expert behavior is further reflected in the latent memories themselves.
On a single training domain, the latent memories produced by different experts at the same prompt occupy clearly separated regions in t-SNE space (Figure~\ref{fig:expert_latents_by_expert}), suggesting that the experts have acquired distinct memory representations rather than collapsing onto one another.
Across stages, the latent memories produced by corresponding experts on math, science, and code prompts also form well-separated clusters (Figure~\ref{fig:expert_latents_cross_domain}), confirming that the stage-isolated expert pools specialize to their corresponding domains rather than producing domain-agnostic outputs.

\paragraph{AE routing analysis.}\label{sec:analysis}

Figure~\ref{fig:tsne_reasoner_hidden} visualizes the hidden states produced by the frozen reasoner at the first latent-memory insertion point for prompts from math, science, and code training data using t-SNE~\citep{VanderMaaten2008Visualizing}.
The separated clusters indicate that prompt representations in the reasoner's hidden-state space provide clear distributional signals for router-group selection and OOD fallback.
This empirical result supports the AE-based routing rule described in Section~\ref{sec:ae_routing}: inputs close to a seen distribution are routed to the corresponding group, while inputs far from all seen distributions fall back to the pretrained model.

Table~\ref{tab:routing_analysis_all_stages} summarizes AE routing statistics across all three CL stages.
At each stage, the prompt-end feature is compared only against the autoencoders from already seen stages; if it is not close to any available stage, the input is rejected as OOD and the model falls back to the pretrained reasoner.
At Stage~1, only the math AE is available, so science and code inputs are mostly rejected, while Math Test is predominantly routed to AE math.
After the science AE is added at Stage~2, Science Test is mostly absorbed by AE science while Math Test remains concentrated on AE math.
After the code AE is added at Stage~3, Code Test is mostly routed to AE code.
This stage-wise progression shows that the router respects the continual-learning setting: it only selects among seen-stage AEs, or chooses OOD when none is compatible.
Across these three datasets, the routing outcome is determined by proximity in the frozen reasoner's hidden-state space rather than by a manually provided task identity.
Before a dataset's stage is learned, its inputs are mostly rejected; once the corresponding AE is added, those inputs are primarily absorbed by that stage.

Importantly, this separation does not come from jointly training the AEs as a domain classifier.
Each AE is trained independently on hidden states from its own domain, and the routing decision is made only by comparing reconstruction errors at inference time.
Thus, the limited cross-domain activation observed in the routing statistics reflects structure already present in the frozen reasoner's representation space, rather than a jointly learned domain boundary.
This property helps protect domains that do not match the current learned distribution: matched inputs can use the corresponding routing group, while unmatched inputs are rejected or fallback, reducing the risk of injecting latent memory from an incompatible domain.
By preserving the pretrained reasoner's behavior on unmatched inputs, this fallback mechanism directly protects the model's original competence and mitigates forgetting under continual adaptation.

%% file: tables/big_table_3iid_default_with_cl_metrics.tex
\begin{table*}[tb]
\centering
\caption{\textbf{Continual learning results under the default order $\text{Math} \rightarrow \text{Science} \rightarrow \text{Code}$.} All methods share Qwen3-4B-Instruct-2507 as the base model. After each training stage we report per-test accuracy and three CL auxiliary metrics. Auxiliary metrics are undefined at $t=1$ and shown as ``--''. \textbf{Bold} marks the best score within each stage. Subscripts on \textbf{Average} mark the gap relative to the Vanilla baseline (\textcolor{blue}{$+$}/\textcolor{red}{$-$}).}
\label{tab:main_3iid_with_cl_metrics}
\renewcommand{\arraystretch}{1.3}
\definecolor{lightblue}{rgb}{0.88, 0.91, 1.00}
\resizebox{\textwidth}{!}{
\begin{tabular}{|l|l|c|c|c|>{\centering\arraybackslash}p{2.0cm}|c|c|c|}
\hline
\multirow{2}{*}{Trained On} & Method & Math Test & Science Test & Code Test & \textbf{Average} ($\uparrow$) & Forget ($\downarrow$) & BWT ($\uparrow$) & FWT ($\uparrow$) \\ \cline{2-9}
& Vanilla & 30.40 & 88.20 & 72.00 & 63.53 & -- & -- & -- \\
\hline
\multirow{4}{*}{\cellcolor{white}\makecell[l]{Stage 1 Math \\ (Nemotron Math)}}
& SFT & 57.20 & 84.20 & 32.80 & 58.07\diffRed{5.46} & -- & -- & -- \\
& ExpeL & 28.00 & 87.00 & 64.20 & 59.73\diffRed{3.80} & -- & -- & -- \\
& MemGen & 58.20 & 87.00 & 40.00 & 61.73\diffRed{1.80} & -- & -- & -- \\
\rowcolor{lightblue} \cellcolor{white} & Ours & \textbf{59.20} & \textbf{88.20} & \textbf{72.00} & \textbf{73.13}\diffBlue{9.60} & -- & -- & -- \\
\hline
\multirow{4}{*}{\cellcolor{white}\makecell[l]{Stage 2 Science \\ (Nemotron Science)}}
& SFT & 29.60 & 86.00 & 15.00 & 43.53\diffRed{20.00} & 27.60 & -27.60 & -4.00 \\
& ExpeL & 31.40 & \textbf{90.80} & 65.00 & 62.40\diffRed{1.13} & \textbf{-3.40} & \textbf{3.40} & -1.20 \\
& MemGen & 33.40 & 89.00 & 16.00 & 46.13\diffRed{17.40} & 24.80 & -24.80 & -1.20 \\
\rowcolor{lightblue} \cellcolor{white} & Ours & \textbf{59.20} & 88.40 & \textbf{72.00} & \textbf{73.20}\diffBlue{9.67} & 0.00 & 0.00 & \textbf{0.00} \\
\hline
\multirow{4}{*}{\cellcolor{white}\makecell[l]{Stage 3 Code \\ (KodCode)}}
& SFT & 12.80 & 27.20 & 63.20 & 34.40\diffRed{29.13} & 51.60 & -51.60 & -30.50 \\
& ExpeL & 32.00 & \textbf{91.80} & 60.60 & 61.47\diffRed{2.06} & \textbf{-0.80} & \textbf{2.50} & -4.10 \\
& MemGen & 32.00 & 84.80 & 73.60 & 63.47\diffRed{0.06} & 15.20 & -15.20 & -28.60 \\
\rowcolor{lightblue} \cellcolor{white} & Ours & \textbf{59.20} & 88.40 & \textbf{74.20} & \textbf{73.93}\diffBlue{10.40} & 0.00 & 0.00 & \textbf{0.00} \\
\hline
\end{tabular}
}
\end{table*}

%% file: tables/big_table_3iid_ablation_with_cl_metrics.tex
\begin{table*}[tb]
\centering
\caption{\textbf{Ablation study.} \textit{Isolation R\&E} (\checkmark{} = each stage has its own router and experts, $\times$ = a single shared group across all stages) and \textit{MoE} (\checkmark{} = multiple experts, $\times$ = a single expert). }
\label{tab:ablation_3iid_with_cl_metrics}
\renewcommand{\arraystretch}{1.3}
\definecolor{lightblue}{rgb}{0.88, 0.91, 1.00}
\resizebox{\textwidth}{!}{
\begin{tabular}{|l|c|c|c|c|c|c|c|c|c|}
\hline
\multirow{2}{*}{Trained On} & \makecell{Isolation \\ R\&E} & \makecell{MoE} & Math Test & Science Test & Code Test & \textbf{Average} ($\uparrow$) & Forget ($\downarrow$) & BWT ($\uparrow$) & FWT ($\uparrow$) \\ \cline{2-10}
& \multicolumn{2}{c|}{Vanilla} & 30.40 & 88.20 & 72.00 & 63.53 & -- & -- & -- \\
\hline
\multirow{4}{*}{\cellcolor{white}\makecell[l]{Stage 1 Math \\ (Nemotron Math)}}
& $\times$ & $\times$ & 53.80 & 86.40 & 44.00 & 61.40 & -- & -- & -- \\
& $\times$ & \checkmark & \textbf{59.80} & 86.60 & 54.40 & 66.93 & -- & -- & -- \\
& \checkmark & $\times$ & 53.80 & \textbf{88.20} & \textbf{72.00} & 71.33 & -- & -- & -- \\
\rowcolor{lightblue} \cellcolor{white} & \checkmark & \checkmark & 59.20 & \textbf{88.20} & \textbf{72.00} & \textbf{73.13} & -- & -- & -- \\
\hline
\multirow{4}{*}{\cellcolor{white}\makecell[l]{Stage 2 Science \\ (Nemotron Science)}}
& $\times$ & $\times$ & 35.20 & 86.00 & 24.60 & 48.60 & 18.60 & -18.60 & -1.80 \\
& $\times$ & \checkmark & 34.80 & 86.80 & 13.20 & 44.93 & 25.00 & -25.00 & -1.60 \\
& \checkmark & $\times$ & 53.80 & 86.00 & \textbf{72.00} & 70.60 & \textbf{0.00} & \textbf{0.00} & \textbf{0.00} \\
\rowcolor{lightblue} \cellcolor{white} & \checkmark & \checkmark & \textbf{59.20} & \textbf{88.40} & \textbf{72.00} & \textbf{73.20} & \textbf{0.00} & \textbf{0.00} & \textbf{0.00} \\
\hline
\multirow{4}{*}{\cellcolor{white}\makecell[l]{Stage 3 Code \\ (KodCode)}}
& $\times$ & $\times$ & 32.00 & 56.20 & 70.00 & 52.73 & 26.00 & -25.80 & -24.60 \\
& $\times$ & \checkmark & 25.80 & 59.80 & 69.80 & 51.80 & 30.50 & -30.50 & -30.20 \\
& \checkmark & $\times$ & 53.80 & 86.00 & 70.80 & 70.20 & 1.10 & \textbf{0.00} & \textbf{0.00} \\
\rowcolor{lightblue} \cellcolor{white} & \checkmark & \checkmark & \textbf{59.20} & \textbf{88.40} & \textbf{74.20} & \textbf{73.93} & \textbf{0.00} & \textbf{0.00} & \textbf{0.00} \\
\hline
\end{tabular}
}
\end{table*}

%% file: Content/06_conclusion.tex
\section{Conclusion}

This paper proposes \methodname, a generative latent memory framework based on dynamic MoE.
Key-query routing selects and weights relevant experts to construct the injected latent memory.
By keeping the core reasoner frozen and training only the MoE memory module, the framework averts catastrophic forgetting at the architectural level.
Experiments on cross-domain continual-learning sequences validate the effectiveness of the framework.

\textbf{Limitation.} Compared with a purely frozen backbone, our method introduces additional expert, router, and autoencoder modules, which incur extra parameters. We view this as a necessary cost: these modules follow a parameter-efficient fine-tuning (PEFT) design and remain compact relative to the backbone, and they provide capacity isolation across domains so that continual adaptation does not cause cross-domain interference.

\textbf{Broader Impacts.} By enabling agents to internalize new knowledge without erasing previously acquired competence, our framework can improve the long-term reliability and applicability of LLM-based agents in dynamically changing real-world environments. We do not anticipate distinctive negative societal impacts beyond those broadly applicable to LLM-based agents.

%% file: Content/07_appendix.tex
\section{Implementation Details}
\label{app:implementation-details}

For AE-based routing, we first run the frozen reasoner once over each prompt and cache the last-layer hidden state at the final prompt token, which is the prompt-end feature at the latent-memory insertion position.
We then train one small autoencoder per training domain to reconstruct this feature in a fully unsupervised manner.

The AE used is a symmetric MLP with encoder $2560 \rightarrow 512 \rightarrow 128$ and decoder $128 \rightarrow 512 \rightarrow 2560$, using ReLU activations.
Training minimizes the reconstruction objective $\mathcal{L}_{\mathrm{AE}} = \lVert \mathrm{AE}_t(\mathbf{h}) - \mathbf{h} \rVert_2^2$ using the Adam optimizer, learning rate $1 \times 10^{-3}$.
The three training domains each train their own AE independently.

Each stage-specific rejection threshold is calibrated independently on its own validation set to cover 95\% of its validation samples.

The MoE module within each stage contains four LoRA experts, each configured with LoRA rank $16$ and alpha $32$; the key-query router selects the top two experts when invoked and aggregates their latent memories with the router-produced weights.
Matching scores are computed with L2-normalized cosine similarity, and routing weights are produced by a softmax over selected experts.

We weight the load-balance loss by $\lambda = 0.01$.
For latent memory, we inject one memory segment immediately after the prompt and insert at most five additional injections during inference; every injected segment is a fixed-length sequence of eight latent tokens.
We train with the AdamW optimizer, learning rate $1 \times 10^{-5}$, and warmup ratio $0.1$.
For each domain, we use a fixed random seed to sample 3{,}000 training, 100 validation, and 500 test instances. The maximum generation length is 2048 tokens. All training and evaluation are conducted on NVIDIA H200 GPUs.

\section{Analysis}
\label{app:analysis}

\begin{figure}[tb]
\centering
\includegraphics[width=0.6\linewidth]{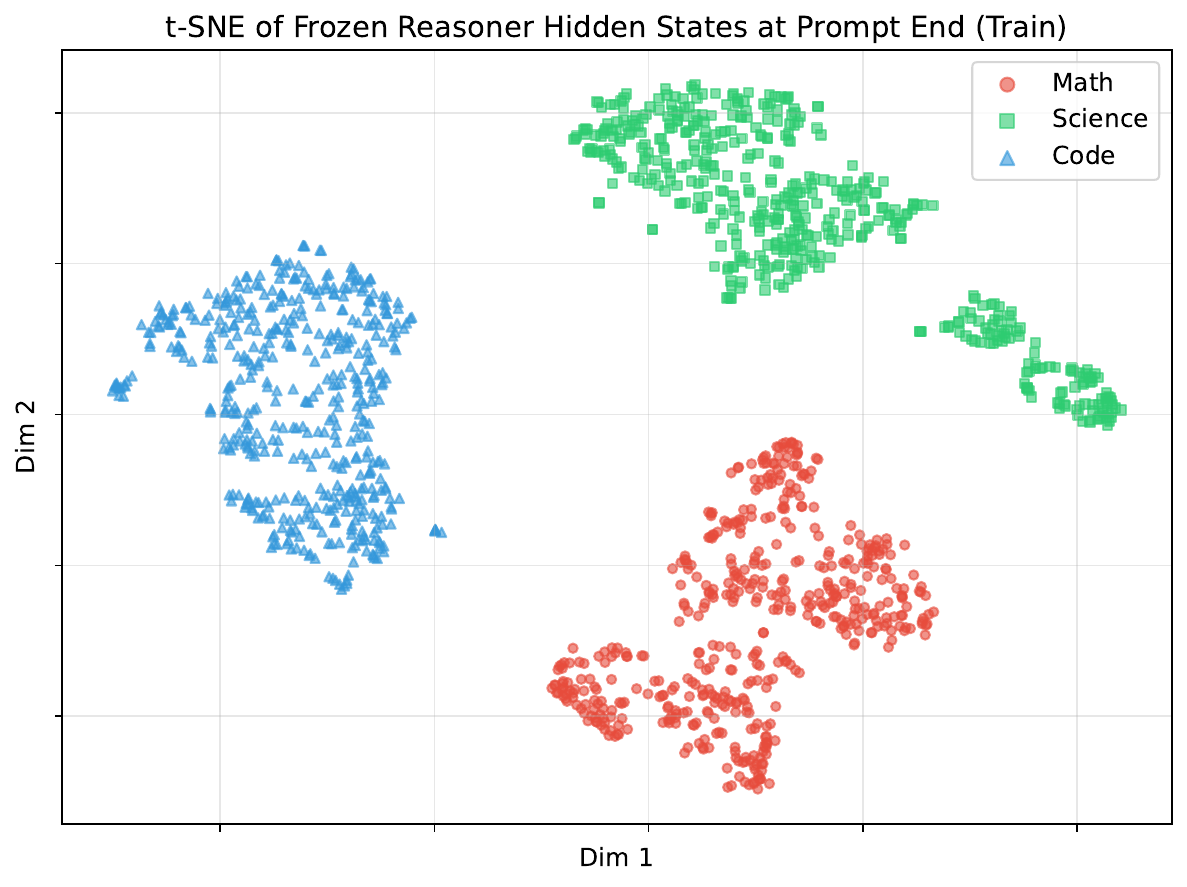}
\caption{t-SNE visualization of frozen-reasoner hidden states at the first latent-memory insertion point for math, science, and code training prompts.}
\label{fig:tsne_reasoner_hidden}
\end{figure}

\paragraph{Hidden-state distributional analysis.}
Figure~\ref{fig:tsne_reasoner_hidden} visualizes the prompt-end hidden states via t-SNE~\citep{VanderMaaten2008Visualizing} for the math, science, and code training prompts, providing the empirical validation for the AE-based routing rule used in Section~\ref{sec:ae_routing}.
The separated clusters indicate that prompt representations in the reasoner's hidden-state space provide distributional signals for router-group selection and OOD fallback.

\paragraph{AE routing statistics.}
Table~\ref{tab:routing_analysis_all_stages} reports the AE routing statistics across continual-learning stages, supporting the analysis in Section~\ref{sec:ae_routing}.
\input{tables/routing_analysis_all_stages}

\begin{figure}[tb]
\centering
\includegraphics[width=0.85\linewidth]{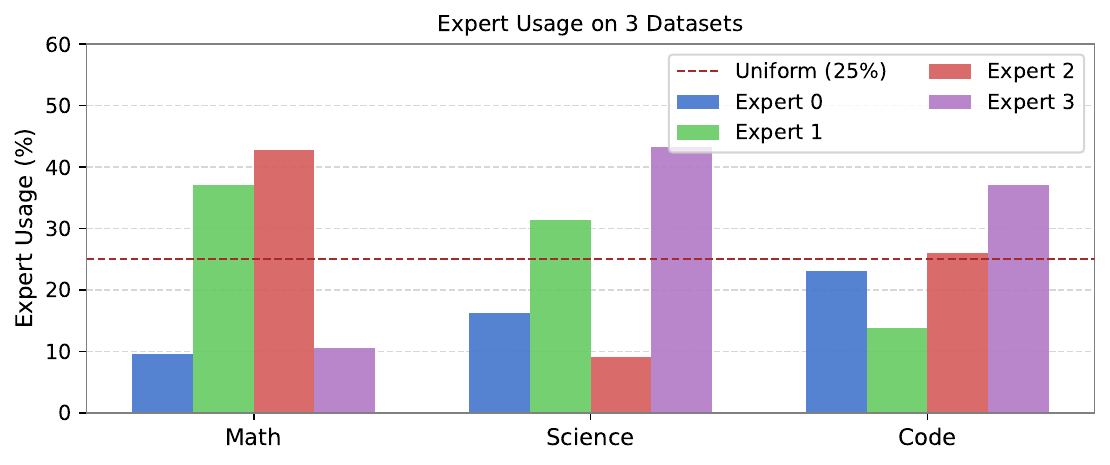}
\caption{Overall expert usage on the Math, Science, and Code test sets.}
\label{fig:expert_usage_3iid_overall}
\end{figure}

\paragraph{Overall expert usage.}
Figure~\ref{fig:expert_usage_3iid_overall} reports the overall expert usage on the Math, Science, and Code test sets, supporting the spontaneous expert specialization discussed in Section~\ref{sec:analysis}.

\begin{figure}[tb]
\centering
\includegraphics[width=0.7\linewidth]{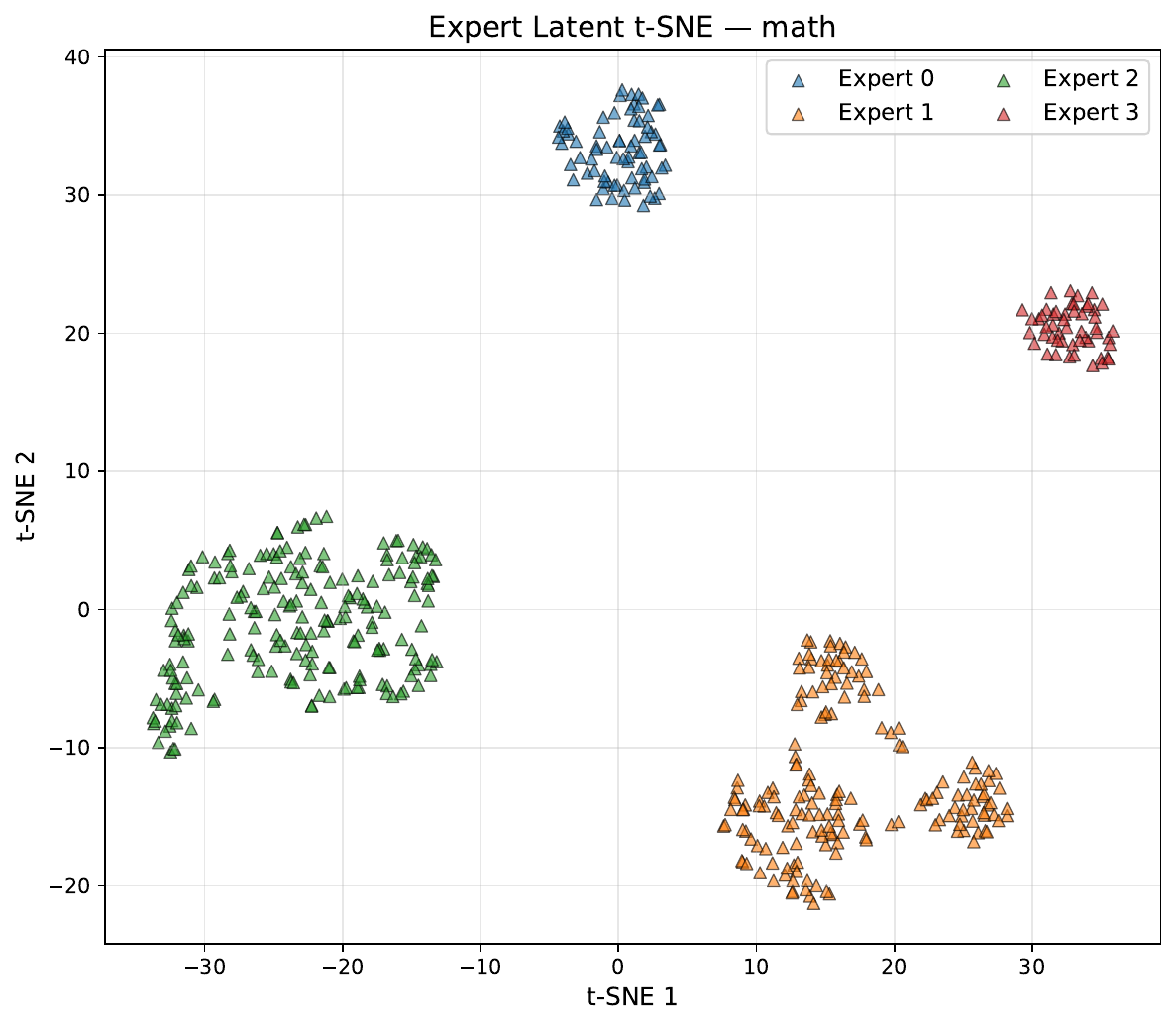}
\caption{t-SNE visualization of latent memories produced by different experts on Math prompts. Different experts occupy distinct regions of the t-SNE space, indicating diverse learned memory representations.}
\label{fig:expert_latents_by_expert}
\end{figure}

\paragraph{Latent memories from different experts on the same domain.}
Figure~\ref{fig:expert_latents_by_expert} visualizes, on Math prompts only, the latent memories produced by each expert via t-SNE.
The latent memories from different experts occupy clearly separated regions of the t-SNE space, suggesting that the experts have learned distinct memory representations.

\begin{figure}[tb]
\centering
\includegraphics[width=0.7\linewidth]{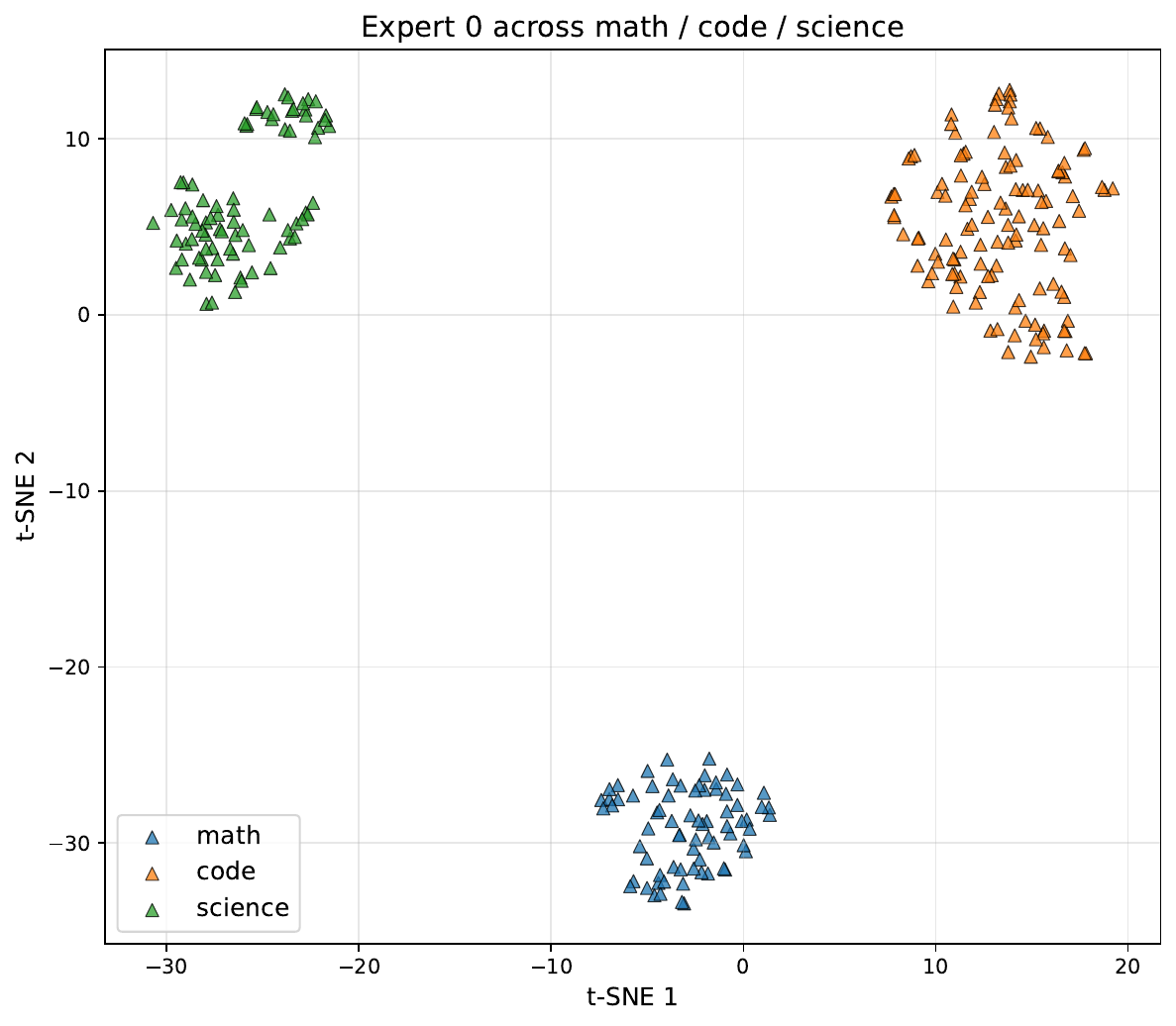}
\caption{t-SNE visualization of latent memories produced by the corresponding expert (within-pool index $0$) on math, science, and code prompts.}
\label{fig:expert_latents_cross_domain}
\end{figure}

\paragraph{Latent memories from different domains.}
Figure~\ref{fig:expert_latents_cross_domain} visualizes the latent memories produced on math, science, and code prompts by the corresponding expert in each stage's pool (within-pool index $0$).
The three stages' latent memories form well-separated clusters, confirming that the stage-isolated expert pools specialize to their corresponding domains rather than producing domain-agnostic outputs.

\section{Additional Results on Training Order}
\label{app:order-sensitivity}

This section reports the results under two alternative training orders to show the effect of task-order variation.
The main paper uses the default order $\text{Math} \rightarrow \text{Science} \rightarrow \text{Code}$, while this appendix additionally reports $\text{Code} \rightarrow \text{Science} \rightarrow \text{Math}$ and $\text{Science} \rightarrow \text{Code} \rightarrow \text{Math}$.
Across all three training orders, our method consistently achieves the highest final Average Performance after the full continual-learning sequence, indicating that its advantage is stable under variations in task ordering.

\input{tables/big_table_3iid_orderA_with_cl_metrics}
\input{tables/big_table_3iid_orderB_with_cl_metrics}

%% file: tables/routing_analysis_all_stages.tex
\begin{table*}[tb]
\centering
\caption{AE routing statistics across continual-learning stages. At each stage, the prompt-end feature is compared only against the autoencoders trained on already-introduced domains; samples far from every available domain AE are rejected as OOD and fall back to the pretrained model.}
\label{tab:routing_analysis_all_stages}
\resizebox{\textwidth}{!}{%
\begin{tabular}{llccc}
\toprule
\textbf{CL Stage} & \textbf{Metric} & \textbf{Math Test} & \textbf{Science Test} & \textbf{Code Test} \\
\midrule
\multirow{2}{*}{\makecell[l]{Stage 1 \\ Math}} & AE math (\%) & 98.00 & 0.00 & 0.00 \\
& OOD (\%) & 2.00 & 100.00 & 100.00 \\
\midrule
\multirow{3}{*}{\makecell[l]{Stage 2 \\ Science}} & AE math (\%) & 98.00 & 0.00 & 0.00 \\
& AE science (\%) & 0.00 & 96.40 & 0.00 \\
& OOD (\%) & 2.00 & 3.60 & 100.00 \\
\midrule
\multirow{4}{*}{\makecell[l]{Stage 3 \\ Code}} & AE math (\%) & 98.00 & 0.00 & 0.00 \\
& AE science (\%) & 0.00 & 96.40 & 0.00 \\
& AE code (\%) & 0.00 & 0.00 & 91.60 \\
& OOD (\%) & 2.00 & 3.60 & 8.40 \\
\bottomrule
\end{tabular}%
}
\end{table*}

%% file: tables/big_table_3iid_orderA_with_cl_metrics.tex
\begin{table*}[tb]
\centering
\caption{\textbf{Continual learning results under Order A: $\text{Code} \rightarrow \text{Science} \rightarrow \text{Math}$.} All methods share Qwen3-4B-Instruct-2507 as the base model. After each training stage we report per-test accuracy and three CL auxiliary metrics. Auxiliary metrics are undefined at $t=1$ and shown as ``--''. \textbf{Bold} marks the best score within each stage. Subscripts on \textbf{Average} mark the gap relative to the Vanilla baseline (\textcolor{blue}{$+$}/\textcolor{red}{$-$}).}
\label{tab:main_3iid_with_cl_metrics_order_A}
\renewcommand{\arraystretch}{1.3}
\definecolor{lightblue}{rgb}{0.88, 0.91, 1.00}
\resizebox{\textwidth}{!}{
\begin{tabular}{|l|l|c|c|c|>{\centering\arraybackslash}p{2.0cm}|c|c|c|}
\hline
\multirow{2}{*}{Trained On} & Method & Code Test & Science Test & Math Test & \textbf{Average} ($\uparrow$) & Forget ($\downarrow$) & BWT ($\uparrow$) & FWT ($\uparrow$) \\ \cline{2-9}
& Vanilla & 72.00 & 88.20 & 30.40 & 63.53 & -- & -- & -- \\
\hline
\multirow{4}{*}{\cellcolor{white}\makecell[l]{Stage 1 Code \\ (KodCode)}}
& SFT & 62.00 & 86.60 & \textbf{50.60} & \textbf{66.40}\diffBlue{2.87} & -- & -- & -- \\
& ExpeL & 62.00 & 89.00 & 30.80 & 60.60\diffRed{2.93} & -- & -- & -- \\
& MemGen & 73.20 & \textbf{89.60} & 33.20 & 65.33\diffBlue{1.80} & -- & -- & -- \\
\rowcolor{lightblue} \cellcolor{white} & Ours & \textbf{74.20} & 88.20 & 30.40 & 64.27\diffBlue{0.74} & -- & -- & -- \\
\hline
\multirow{4}{*}{\cellcolor{white}\makecell[l]{Stage 2 Science \\ (Nemotron Science)}}
& SFT & 59.80 & 86.80 & 27.00 & 57.87\diffRed{5.66} & 2.20 & -2.20 & -1.60 \\
& ExpeL & 58.80 & \textbf{91.80} & 31.40 & 60.67\diffRed{2.86} & 3.20 & -3.20 & 0.80 \\
& MemGen & 6.80 & 90.60 & \textbf{35.40} & 44.27\diffRed{19.26} & 66.40 & -66.40 & \textbf{1.40} \\
\rowcolor{lightblue} \cellcolor{white} & Ours & \textbf{74.20} & 88.40 & 30.80 & \textbf{64.47}\diffBlue{0.94} & \textbf{0.00} & \textbf{0.00} & 0.00 \\
\hline
\multirow{4}{*}{\cellcolor{white}\makecell[l]{Stage 3 Math \\ (Nemotron Math)}}
& SFT & 59.80 & 73.20 & 54.80 & 62.60\diffRed{0.93} & 7.90 & -7.90 & -2.50 \\
& ExpeL & 60.40 & \textbf{91.80} & 29.60 & 60.60\diffRed{2.93} & 0.80 & -0.80 & 0.90 \\
& MemGen & 29.60 & 85.00 & 58.40 & 57.67\diffRed{5.86} & 24.60 & -24.60 & \textbf{3.20} \\
\rowcolor{lightblue} \cellcolor{white} & Ours & \textbf{74.20} & 88.40 & \textbf{59.20} & \textbf{73.93}\diffBlue{10.40} & \textbf{0.00} & \textbf{0.00} & 0.20 \\
\hline
\end{tabular}
}
\end{table*}

%% file: tables/big_table_3iid_orderB_with_cl_metrics.tex
\begin{table*}[tb]
\centering
\caption{\textbf{Continual learning results under Order B: $\text{Science} \rightarrow \text{Code} \rightarrow \text{Math}$.} All methods share Qwen3-4B-Instruct-2507 as the base model. After each training stage we report per-test accuracy and three CL auxiliary metrics. Auxiliary metrics are undefined at $t=1$ and shown as ``--''. \textbf{Bold} marks the best score within each stage. Subscripts on \textbf{Average} mark the gap relative to the Vanilla baseline (\textcolor{blue}{$+$}/\textcolor{red}{$-$}).}
\label{tab:main_3iid_with_cl_metrics_order_B}
\renewcommand{\arraystretch}{1.3}
\definecolor{lightblue}{rgb}{0.88, 0.91, 1.00}
\resizebox{\textwidth}{!}{
\begin{tabular}{|l|l|c|c|c|>{\centering\arraybackslash}p{2.0cm}|c|c|c|}
\hline
\multirow{2}{*}{Trained On} & Method & Science Test & Code Test & Math Test & \textbf{Average} ($\uparrow$) & Forget ($\downarrow$) & BWT ($\uparrow$) & FWT ($\uparrow$) \\ \cline{2-9}
& Vanilla & 88.20 & 72.00 & 30.40 & 63.53 & -- & -- & -- \\
\hline
\multirow{4}{*}{\cellcolor{white}\makecell[l]{Stage 1 Science \\ (Nemotron Science)}}
& SFT & 86.60 & 27.20 & 31.20 & 48.33\diffRed{15.20} & -- & -- & -- \\
& ExpeL & \textbf{90.60} & 65.40 & 32.00 & 62.67\diffRed{0.86} & -- & -- & -- \\
& MemGen & 90.00 & 62.20 & \textbf{34.00} & 62.07\diffRed{1.46} & -- & -- & -- \\
\rowcolor{lightblue} \cellcolor{white} & Ours & 88.40 & \textbf{72.00} & 30.80 & \textbf{63.73}\diffBlue{0.20} & -- & -- & -- \\
\hline
\multirow{4}{*}{\cellcolor{white}\makecell[l]{Stage 2 Code \\ (KodCode)}}
& SFT & 62.00 & 59.20 & 20.20 & 47.13\diffRed{16.40} & 24.60 & -24.60 & -44.80 \\
& ExpeL & \textbf{91.00} & 58.60 & 30.20 & 59.93\diffRed{3.60} & \textbf{-0.40} & \textbf{0.40} & -6.60 \\
& MemGen & 87.40 & 73.40 & 23.00 & 61.27\diffRed{2.26} & 2.60 & -2.60 & -9.80 \\
\rowcolor{lightblue} \cellcolor{white} & Ours & 88.40 & \textbf{74.20} & \textbf{30.80} & \textbf{64.47}\diffBlue{0.94} & 0.00 & 0.00 & \textbf{0.00} \\
\hline
\multirow{4}{*}{\cellcolor{white}\makecell[l]{Stage 3 Math \\ (Nemotron Math)}}
& SFT & 73.60 & 59.40 & 54.80 & 62.60\diffRed{0.93} & 6.40 & -6.40 & -27.50 \\
& ExpeL & \textbf{91.20} & 60.20 & 31.20 & 60.87\diffRed{2.66} & 2.50 & \textbf{1.10} & -3.40 \\
& MemGen & 84.40 & 68.60 & 56.20 & 69.73\diffBlue{6.20} & 5.20 & -5.20 & -8.60 \\
\rowcolor{lightblue} \cellcolor{white} & Ours & 88.40 & \textbf{74.20} & \textbf{59.20} & \textbf{73.93}\diffBlue{10.40} & \textbf{0.00} & 0.00 & \textbf{0.20} \\
\hline
\end{tabular}
}
\end{table*}